\documentclass{article}

\PassOptionsToPackage{numbers,sort&compress}{natbib}
\PassOptionsToPackage{table}{xcolor}
\usepackage[utf8]{inputenc}
\usepackage[T1]{fontenc}
\usepackage{iclr2027_conference,times}
\iclrfinalcopy

\usepackage{amsmath,amsfonts,amssymb,amsthm,mathtools,bm}
\usepackage{algorithm,algorithmic,float}
\usepackage{array,booktabs,makecell,multirow}
\usepackage{graphicx,subcaption,wrapfig}
\usepackage{enumitem,microtype,nicefrac,relsize,soul,verbatim,xcolor,xspace}
\usepackage{url}
\usepackage[hidelinks]{hyperref}

\newcommand{\locfg}{\textsc{LocFG}\xspace}
\newcommand{\locar}{\textsc{LoCaR}\xspace}

\newcommand{\supportpair}[2]{%
  \begin{tabular}[c]{@{}l@{}}
    \textbf{Tr}: #1\\
    \textbf{ID}: #2
  \end{tabular}%
}

\definecolor{bestgreen}{RGB}{0,110,0}
\definecolor{secondgreen}{RGB}{190,220,190}
\newcommand{\best}[1]{\cellcolor{bestgreen}\textcolor{white}{\textbf{#1}}}
\newcommand{\secondbest}[1]{\cellcolor{secondgreen}\textbf{#1}}

\theoremstyle{definition}

\theoremstyle{remark}

\newcommand{\ashirbad}[1]{}
\newcommand{\Dipanwita}[1]{}
\newcommand{\Rahim}[1]{}

\newcommand{\PaperTitle}{LoCaR: Localized Canvas Retrieval with Frozen Vision-Language Models}
\title{\raggedright\PaperTitle}

\author{%
  \textbf{Rahim Hossain}\textsuperscript{2,\(\dagger\)}\quad
  \textbf{Dipanwita Saha}\textsuperscript{1}\quad
  \textbf{Ashirbad Mishra}\textsuperscript{1}\\
  \textbf{Xinxin Shu}\textsuperscript{1}\quad
  \textbf{Svetlana Strunjas}\textsuperscript{1}\quad
  \textbf{Binbin Li}\textsuperscript{1}\\[0.5em]
  \textsuperscript{1}eBay Inc.\qquad
  \textsuperscript{2}Binghamton University\\[0.25em]
  \href{mailto:rhossain@binghamton.edu}{\small\texttt{rhossain@binghamton.edu}}\\
  {\small\texttt{\{dipsaha,ashirmishra,xshu,sstrunjas,binbli\}@ebay.com}}\\[0.35em]
  {\small\textsuperscript{\(\dagger\)}Work done while Rahim Hossain was a research intern at eBay Inc.}%
}

\hypersetup{
  pdftitle={\PaperTitle},
  pdfauthor={Rahim Hossain, Dipanwita Saha, Ashirbad Mishra, Xinxin Shu, Svetlana Strunjas, Binbin Li},
  pdfsubject={Localized text-to-image retrieval with frozen vision-language models}
}

\begin{document}
\maketitle
\lhead{\small Preprint}

\begin{abstract}
Whole-image embeddings can under represent small objects and local attributes that distinguish relevant images for localized queries. We present LoCaR, a training-free framework for localized text-to-image retrieval with frozen vision–language models. LoCaR constructs and utilizes independent proposals (SAM2) focusing on the object so as to get a more localized representation that independently encodes each region using the frozen visual encoder. We further study paired-support adaptation as an optional residual that uses matched text–region exemplars without updating the underlying encoders. We also introduce \locfg{}, a Visual Genome-derived benchmark designed to evaluate localized retrieval for smaller and more challenging referents.
Across several vision language backbones, our results show that improving the region representation itself is the main source of gain. 
\end{abstract}
\section{Introduction}
\label{sec:introduction}

Image search queries often describe an object or attribute within a larger
scene, such as ``a woman carrying a yellow handbag.'' Retrieving the
relevant image requires preserving local evidence that may occupy only
a small part of the photograph. Vision--language models such as CLIP
and SigLIP2 provide efficient retrieval through aligned image and text
embeddings, but a single whole-image representation can underrepresent
these details~\citep{radford2021learning,tschannen2025siglip2}.
We study \emph{localized image retrieval}, where a referring expression
is used to rank images across a gallery. This differs from
referring-expression comprehension, which localizes an object within
an already supplied image.

Region representations offer a way to retain local evidence until a
query is available. TextRegion constructs text-aligned region embeddings
from segmentation masks and frozen image--text models, providing a
strong foundation for localized matching~\citep{xiao2025textregion}.
Its region vectors aggregate features extracted from the surrounding
image. An alternative is to construct a region-specific visual input
and encode it independently. Cropping and background suppression have
established precedents in ReCLIP  and fine-grained visual
prompting~\citep{subramanian2022reclip,yang2023fgvp}.
We investigate how this alternative representation affects gallery
retrieval under matched backbones and candidate proposals.

We introduce \locar{}, a frozen region-canvas retrieval framework.
Each SAM2 proposal~\citep{ravi2024sam2} defines a padded crop in which
foreground pixels are preserved and the background is blurred.
The frozen vision--language model independently encodes each canvas
using its standard image readout. Gallery representations are cached
offline, and the maximum query--region similarity supplies an
image-level retrieval score. A whole-image similarity channel provides
complementary scene context. Within each backbone, comparisons with
TextRegion ~\citep{xiao2025textregion} preserve proposal identities, query embeddings, and global
embeddings, measuring the combined effect of cropping, background
treatment, and independent encoding.

We separately investigate paired-support adaptation through an optional residual cache. A query and candidate region are compared
through their affinity profiles over matched support expressions and
visual anchors. Agreement over corresponding support examples
contributes additional matching evidence without updating the encoders.
Applying this residual to both TextRegion and canvas representations
separates its incremental benefit from the representation change.


We show experiments on the dataset RefCOCOg ~\citep{mao2016refcocog}. Our method \locar{} improves unadapted
region R@1 on multiple Recall@k on several frozen vision backbones which are CLIP-B/16, SigLIP2-B/16 and Qwen3-VL-Embedding-2B. Our method performs significantly better with  2.40-6.05pp for CLIP, 3.13-5.94 pp for SigLIP2, and 9.13-11.69 pp for Qwen wrt zero shot backbones and beats the SOTA zero-shot retriever TextRegion on all the backbones. We show in the results that the strongest improvement comes from the local region retrieval where the attention only adheres to each region by constructing independent region canvases which enables the attention to be localized. We introduce \locfg{} a Visual Genome-derived benchmark designed for smaller referents~\citep{krishna2017visualgenome}.  We further evaluate \locar{} on \textsc{LocFG}  with experiments using both transferred and in-domain support. We also adapt our method to recent backbone such as Qwen3-VL-Embedding-2B to show that \locar{} can extend to specific scenarios.


Our contributions are:
\begin{enumerate}
    \item We introduce \locar{}, a training-free localized retrieval
    framework that converts segmentation proposals into independently
    encoded region canvases. The resulting embeddings are cached offline
    and can be combined with whole image similarity at retrieval time
    while keeping the vision--language model fully frozen.

    \item We show through controlled comparisons and ablations that
    \textbf{region representation is the main source of improvement}.
    In particular, canvas construction and the text-aligned image readout
    have a much larger effect than paired-support adaptation which
    provides a smaller and setting-dependent correction.

    \item We introduce \textsc{LocFG}, a small-referent retrieval benchmark,
    and use RefCOCOg and \textsc{LocFG} to study how region representation,
    support domain and global--local fusion affect localized retrieval
    across CLIP, SigLIP2 and Qwen backbones.
\end{enumerate}
\section{Related work}
\label{sec:related}
\vspace{-10pt}
\paragraph{Frozen region representations and visual prompting.}
TextRegion~\citep{xiao2025textregion} constructs text-aligned region
tokens by aggregating frozen visual features under SAM2 masks,
providing our principal representation baseline.
An alternative is to modify the visual input before encoding.
ReCLIP~\citep{subramanian2022reclip} scores object proposals using
cropped and background-blurred views with frozen CLIP.
Fine-Grained Visual Prompting
(FGVP)~\citep{yang2023fgvp} uses SAM masks to preserve a target's
foreground while blurring the surrounding background on the full
image. These methods establish cropping and foreground-preserving
background suppression as useful visual prompts.
\locar{} combines padded proposal crops with mask-guided background
blur and independently encodes the resulting canvases for gallery
retrieval. Our matched comparison with TextRegion measures the
combined representation change while holding the backbone and
candidate proposals fixed.


\paragraph{Support-based adaptation.}
Tip-Adapter~\citep{zhang2022tipadapter} augments zero-shot
CLIP~\citep{radford2021learning} classification scores with a cache
of support-image features and one-hot class labels.
CLIP-Adapter~\citep{gao2024clipadapter} learns lightweight feature
adapters, CoOp and CoCoOp~\citep{zhou2022coop,zhou2022cocoop}
optimize text prompts, and APE~\citep{zhu2023ape} refines
pretrained features for few-shot classification.
Our optional paired-support residual uses matched expression--region
anchors rather than class-label values.
Text and visual affinity profiles retain corresponding support-pair
indices, allowing their agreement to contribute to retrieval scores.
This construction accommodates free-form queries without specifying
a fixed classification vocabulary.
The encoders remain frozen, and setting the residual weight to zero
recovers the corresponding base region scorer.
We evaluate the residual on both TextRegion and canvas representations
to distinguish support adaptation from representation changes.

\paragraph{Localized retrieval benchmarks.}
Referring-expression datasets such as RefCOCO and
RefCOCOg~\citep{yu2016refcoco,mao2016refcocog} associate descriptions
with objects inside individual images.
REIR~\citep{reir2025} extends this setting to retrieval across images
with instance localization.
SORCE~\citep{liu2025sorce} evaluates image retrieval from small-object
descriptions and studies both frozen regional prompting and
fine-tuned embeddings.
ILIAS~\citep{kordopatis2025ilias} evaluates instance-level retrieval
against large distractor galleries.
Our evaluation ranks gallery images from localized descriptions and
measures image-retrieval recall, without requiring a predicted box
to satisfy a localization threshold.
We additionally construct \textsc{LocFG} from Visual
Genome~\citep{krishna2017visualgenome} to examine retrieval for
smaller referents, with image-disjoint fit and test partitions
and transferred versus in-domain support.
\section{Region-Canvas Retrieval}
\label{sec:canvas-method}
\begin{figure}[t]
    \centering
    \includegraphics[width=\linewidth]
        {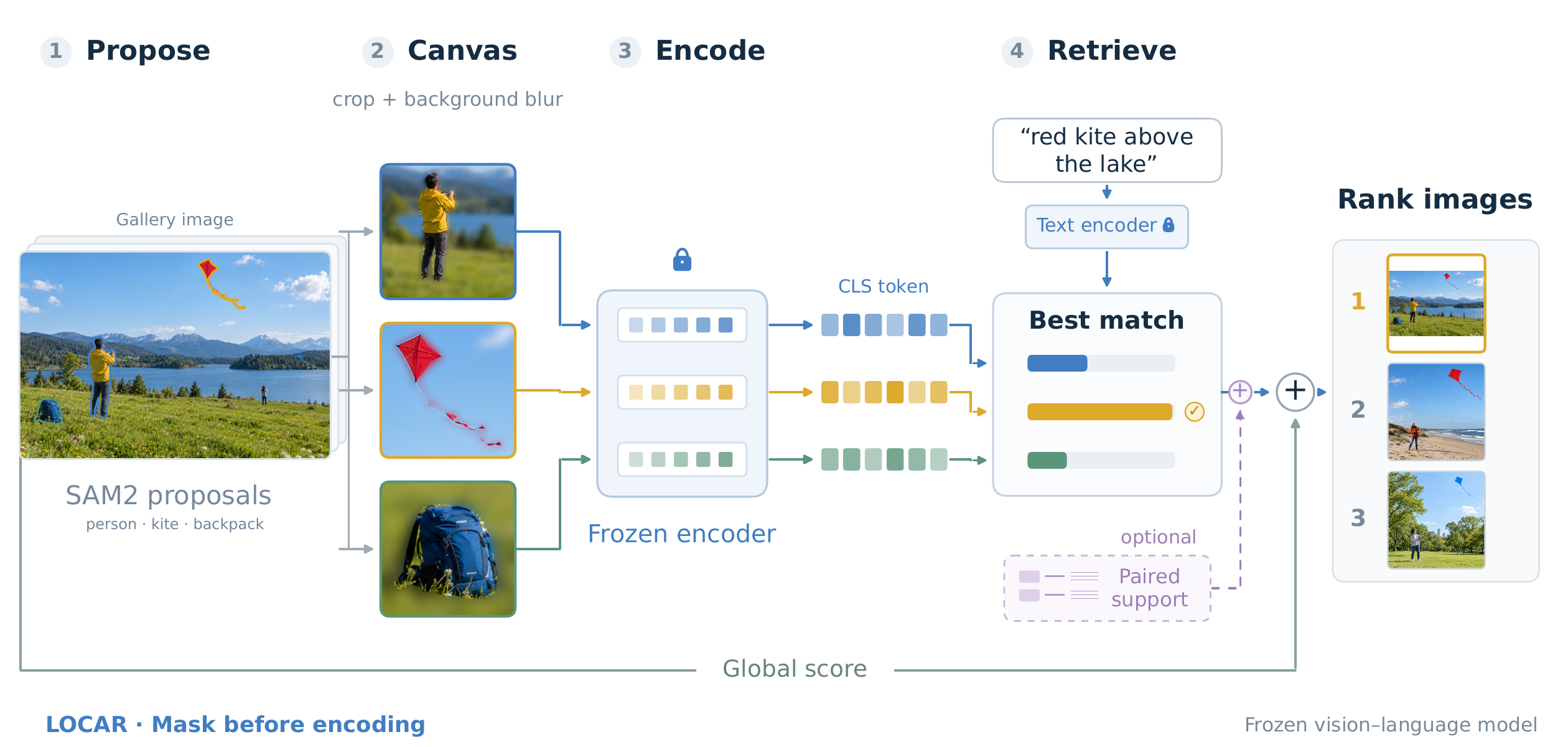}
    \caption{
        \textbf{Region-canvas retrieval.}
        (1) SAM2 generates candidate regions.
        (2) Each proposal defines a foreground-preserving crop with
        blurred background.
        (3) A shared frozen encoder independently embeds each canvas,
        and the embeddings are cached.
        (4) Query--region matching, optional paired-support evidence,
        and whole-image similarity produce the image ranking.
    }
    \label{fig:canvas-overview}
\end{figure}

\vspace{-10pt}
Figure~\ref{fig:canvas-overview} summarizes our pipeline:
generate proposals, construct region canvases, encode them with a
frozen vision--language model, and retrieve images using cached
representations. Each image receives a maximum query--region
similarity score, optionally augmented by paired-support evidence
and combined with whole-image similarity. The vision--language
encoders remain frozen throughout.

\subsection{Task and proposal cache}

Let $\mathcal{G}$ denote the image gallery and $T$ a query describing
a referent. Let $f_t$ and $f_v$ be the text and image embedding
functions of a fixed vision--language model. We define
$\operatorname{nrm}(x)=x/\lVert x\rVert_2$ and obtain the normalized
query embedding as $t=\operatorname{nrm}(f_t(T))$.

SAM2~\citep{ravi2024sam2} generates class-agnostic masks and bounding
boxes for each gallery image. Let $N_I$ denote the number of cached
SAM2 proposals available for image $I$, before applying the scoring
cap described below. We reuse the same proposal identities and
ordering for TextRegion and canvas representations. Proposals may
overlap or represent an object and its parts; they need not partition
the image. Gallery proposals and embeddings are constructed without
query annotations or ground-truth referent boxes.

\subsection{Constructing and encoding region canvases}
\label{sec:canvas-encoding}

Stages~2--3 of Figure~\ref{fig:canvas-overview} construct and encode
a separate visual input for each proposal.
We map the proposal box from the cached mask coordinate frame to
the source image, expand it by $10\%$ of its width and height on
each side, enforce a minimum side length of eight pixels, and clip
it to the image boundary.
Let $J_i$ denote the resulting RGB crop and $\widetilde m_i$ its
aligned binary mask.
Cached masks are stored within their tight proposal boxes at a
maximum long-side resolution of 64 pixels. We reconstruct each mask
by nearest-neighbor resizing into its proposal footprint and assign
zero mask values to the added crop margin.

The canvas preserves foreground pixels and blurs the surrounding
background:
\begin{equation}
C_i
=
\widetilde m_i \odot J_i
+
(1-\widetilde m_i)\odot\mathcal{B}_{55}(J_i),
\label{eq:canvas-composite}
\end{equation}
where $\odot$ broadcasts the mask over RGB channels and
$\mathcal{B}_{55}$ denotes Gaussian blur with a $55\times55$ kernel.
We use OpenCV's kernel-derived default standard deviation and apply
blurring at the crop's original resolution, before resizing the
composite to $224\times224$ for CLIP-B/16 and SigLIP2-B/16.
The blurred background retains coarse context while suppressing
detail outside the foreground mask.

Let $P$ denote resizing followed by the backbone's image
normalization. The region embedding is
\begin{equation}
r_i
=
\operatorname{nrm}\!\left(f_v(P(C_i))\right).
\label{eq:canvas-embedding}
\end{equation}

We use the model's standard text-aligned image readout:
the projected class token for CLIP-B/16 and SigLIP2-B/16.
Canvases are encoded independently, with batching permitted.
Thus, visual attention operates on each constructed canvas before
a region vector is produced.

For Qwen3-VL-Embedding-2B, we reuse the cached SAM2 proposal boxes
and independently encode each plain RGB crop through the model's
image-embedding interface, without background suppression.
This avoids requiring a patch-to-text projection for mask pooling.
The normalized crop embeddings use the same retrieval and
paired-support scoring as the canvas embeddings.
For in-domain \locfg{} evaluation, the query instruction
``Find the photo containing the part of the scene that this phrase
describes'' is selected on fit data before test evaluation
(Appendix~\ref{app:qwen}).

We cache proposal embeddings in their original order and retain
the first $R_I=\min(N_I,32)$ candidates for scoring:
\begin{equation}
\mathcal{R}(I)=\{r_i\}_{i=1}^{R_I}.
\label{eq:region-set}
\end{equation}
Here, $N_I$ counts all available cached proposals for image $I$,
whereas $R_I$ counts those used by the retrieval scorer.
The cache may therefore contain more embeddings than are scored.
Gallery candidates are constructed independently of support matching.
Independent encoding increases offline visual computation relative
to pooling shared image features; the cached embeddings are reused
for all queries.

\subsection{Region-based image scoring}

The retrieval stage in Figure~\ref{fig:canvas-overview} compares the
query with every retained region and assigns each image its largest
query--region cosine similarity:
\begin{equation}
s_0(T,I)
=
\max_{r\in\mathcal{R}(I)}\langle t,r\rangle.
\label{eq:base-region-score}
\end{equation}
We rank gallery images by this score.
Regions supply evidence for image retrieval; the relevant image
is not provided in advance.
The base scorer requires no support examples.
Its effectiveness depends on the retained proposals: a missing or
poorly framed referent may not be represented adequately by any
canvas.

\subsection{Optional paired-support residual}
\label{sec:canvas-support}

We optionally add a \locar{}+ residual inspired by
support-cache adaptation~\citep{zhang2022tipadapter}.
For each labeled support expression, we match its ground-truth box
to the retained proposal with the highest box IoU after aligning
their coordinate frames.
We discard matches below the threshold $\tau_{\mathrm{IoU}}$.
The normalized expression embedding and matched visual embedding
form a support pair $(u_j,v_j)$ in
$\mathcal{B}=\{(u_j,v_j)\}_{j=1}^{M}$.
We retain at most $4{,}096$ pairs, subsampling with a fixed seed
when necessary.
Visual anchors use the same backbone and region representation
as the gallery candidates.

For $\beta>0$, affinities to the text and visual anchors are
\begin{equation}
\begin{aligned}
a_t(t)_j
&=
\exp\!\left[\beta(\langle t,u_j\rangle-1)\right],\\
a_r(r)_j
&=
\exp\!\left[\beta(\langle r,v_j\rangle-1)\right].
\end{aligned}
\label{eq:anchor-affinities}
\end{equation}
We sparsify and normalize these affinities:
\begin{equation}
\begin{aligned}
\phi_t(t)
&=
\operatorname{nrm}\!\left(\operatorname{TopK}(a_t(t))\right),\\
\phi_r(r)
&=
\operatorname{nrm}\!\left(\operatorname{TopK}(a_r(r))\right).
\end{aligned}
\label{eq:anchor-profiles}
\end{equation}
The $\operatorname{TopK}$ operator retains the $K$ largest entries
at their original indices and sets all others to zero.
Both profiles therefore use the same $M$ support-pair coordinates,
and their inner product measures agreement over matched exemplars.

The adapted score is
\begin{equation}
s_{\gamma}(T,I)
=
s_0(T,I)
+
\gamma
\max_{r\in\mathcal{R}(I)}
\langle\phi_t(t),\phi_r(r)\rangle,
\qquad \gamma\geq0.
\label{eq:score}
\end{equation}
The direct-similarity and support-agreement terms are maximized
independently and may select different proposals.
Setting $\gamma=0$ exactly recovers the base region scorer.
Applying the same residual to TextRegion embeddings gives the
TextRegion~+ configuration evaluated in
Sec.~\ref{sec:canvas-ablation}.

\subsection{Global fusion and calibration}
\label{sec:canvas-fusion}

Let $g(I)$ be the normalized whole-image embedding and define
$s_{\mathrm{glob}}(T,I)=\langle t,g(I)\rangle$.
Within each backbone, representation comparisons reuse the same
cached text and whole-image embeddings; canvas construction changes
only the region vectors.
For a fixed query, $z_T[s](I)$ standardizes channel $s$ using its
mean and standard deviation across the gallery.
The fused score is
\begin{equation}
s_{\mathrm{fuse}}(T,I)
=
z_T[s_{\mathrm{glob}}](I)
+
\alpha\,z_T[s_{\gamma}](I).
\label{eq:fusion}
\end{equation}

Support images are disjoint from the evaluation gallery.
For calibration, we divide support images into fit and held-out
halves and use fit anchors to score held-out queries and images.
Affinity sharpness $\beta$ is derived from support similarity
statistics. We select $K$ and $\gamma$ using held-out retrieval
performance, then select $\alpha$ conditional on that residual
configuration.
The base and adapted rows within each comparison share this
$\alpha$, measuring the residual's effect at a fixed fusion weight.
All selected settings are fixed before test evaluation.

Transferred evaluation retains the RefCOCOg support bank and
calibration. In-domain evaluation constructs the bank and selects
hyperparameters using the \textsc{LocFG} fit partition.
Consequently, unadapted fused scores can differ across settings
through $\alpha$.
Unadapted region scores remain identical when query instructions
and visual representations are fixed; Qwen's settings additionally
differ in their query instructions..
\providecommand{\locfg}{\textsc{LocFG}}
\providecommand{\best}[1]{\textbf{#1}}
\providecommand{\secondbest}[1]{\underline{#1}}
\providecommand{\supportpair}[2]{}
\renewcommand{\supportpair}[2]{%
  \begin{tabular}[c]{@{}l@{}}\textbf{Tr}: #1\\\textbf{ID}: #2\end{tabular}%
}

\section{Evaluation}
\label{sec:eval}
\label{sec:results}
\vspace{-10pt}

\subsection{Experimental setup}
\label{sec:setup}
\vspace{-10pt}

We evaluate image retrieval on RefCOCOg~\citep{mao2016refcocog}
($9{,}602$ queries, $2{,}600$ gallery images) and introduce \locfg{},
a Visual Genome-derived benchmark~\citep{krishna2017visualgenome}
for retrieval from descriptions of small referents.
RefCOCOg provides an established evaluation setting, but our audit
of the public mirror revealed split-resolution and validation test
overlap issues can contaminate support-set construction when provenance is inferred only from nominal split labels
(Appendix~\ref{sec:contamination}).
\locfg{} complements this evaluation in two ways: it tests retrieval
when the described object occupies little of the image, and it
provides independently constructed image-disjoint fit/test
partitions with overlap filtering against RefCOCOg.
It contains $12{,}033$ test queries over $2{,}600$ gallery images;
the median referent bounding box occupies $3.0\%$ of image area,
compared with $14.5\%$ on RefCOCOg.
This setting tests whether region canvases preserve evidence about
small targets and whether paired-support gains generalize beyond
RefCOCOg, using either transferred RefCOCOg support or in-domain
\locfg{} support. Construction and filtering details appear in
Appendix~\ref{app:locfg-stress}.

We use frozen CLIP-B/16~\citep{radford2021learning},
SigLIP2-B/16~\citep{tschannen2025siglip2}, and
Qwen3-VL-Embedding-2B~\citep{qwen3vl2025} encoders as backbones for
evaluating \locar{}. Matched-representation comparisons reuse the same
SAM2~\citep{ravi2024sam2} proposals, and hyperparameters are selected on
held-out support images before test evaluation. Details of benchmark
construction, support filtering, proposal settings, and calibration are
provided in Appendix~\ref{app:eval-protocol}; the support-set
contamination audit is reported in Appendix~\ref{sec:contamination}.

\subsection{Representations, baselines, and metrics}
\label{sec:baselines-metrics}
\vspace{-10pt}

We use \emph{TextRegion}~\citep{xiao2025textregion} as our strongest
directly comparable training-free region baseline.
Its published CLIP configuration surpasses the reported zero-shot
RefCOCOg referring-expression comprehension results of
RedCircle~\citep{shtedritski2023redcircle},
FGVP~\citep{yang2023fgvp}, and CPT~\citep{yao2024cpt}.
TextRegion constructs text-aligned region embeddings by pooling
patch value features with proposal-mask weights through the frozen
backbone's readout.
Here, we evaluate these representations for gallery-level image
retrieval. Comparisons with \locar{} share the backbone, proposals,
query embeddings, and whole-image embeddings, while changing how
each proposal is represented.

\emph{\locar{}} independently encodes each proposal as a canvas
that preserves the foreground and blurs the surrounding background
(Appendix~\ref{app:canvas-construction}).
For Qwen, we use plain proposal crops because its embedding interface
does not expose the patch-to-text projection required by our
TextRegion implementation.
``+'' denotes paired expression region support:
\emph{TextRegion +} applies the residual to mask-pooled features,
whereas \emph{\locar{} +} applies it to independently encoded regions.

\emph{Global zero-shot} ranks gallery images using whole-image
cosine similarity to the query.
\emph{Region} uses the local image score
$s_{\gamma}(T,I)$ from Eq.~\ref{eq:score}: the maximum
query-region cosine similarity, optionally augmented by the
paired-support residual.
Without support, $\gamma=0$ and this reduces to
$s_0(T,I)$ in Eq.~\ref{eq:base-region-score}.
\emph{Fused} combines the global and local channels after
per-query z-score standardization: each channel's mean across
gallery images is subtracted and the result is divided by its
standard deviation (Eq.~\ref{eq:fusion}).
Each matched pair of results without and with support uses the
same fusion weight within its representation and support setting.

We report R@1, R@5, and R@10 (\%), the percentages of queries
whose source image appears among the top 1, 5, and 10 retrieved
gallery images, respectively.
Gains are expressed in percentage points (pp), unless stated
otherwise.
Evaluation and calibration details, including the distinction
between standalone and matched global channels, appear in
Appendices~\ref{app:eval-protocol} and~\ref{app:zero-shot}.

\subsection{Main results on RefCOCOg}
\label{sec:canvas-refcocog}
\vspace{-10pt}

\begin{table}[t]
\centering
\caption{\textbf{RefCOCOg retrieval} (\%).
The right-hand columns report standalone global recall for
zero-shot rows and global--region fusion otherwise.
``+'' adds paired support.
Best and second-best values are marked within each backbone
and metric. Standalone and matched fusion global channels are
compared in Appendix~\ref{app:zero-shot}.}
\label{tab:main-results}
\label{tab:full-refcocog}
\label{tab:canvas-ablation}

\begin{tabular}{@{}llrrrrrr@{}}
\toprule
& & \multicolumn{3}{c}{\textbf{Region retrieval}}
  & \multicolumn{3}{c}{\textbf{Global / fused}} \\
\cmidrule(lr){3-5}
\cmidrule(l){6-8}
\textbf{Configuration} & \textbf{Model}
  & R@1 & R@5 & R@10
  & R@1 & R@5 & R@10 \\
\midrule

Zero-shot (Global) & \multirow{5}{*}{CLIP-B/16}
  & -- & -- & --
  & 15.15 & 32.83 & 42.89 \\

TextRegion &
  & 9.66 & 26.25 & 37.94
  & 17.44 & 37.40 & 48.33 \\

TextRegion + &
  & 10.42
  & \secondbest{27.88}
  & \best{39.71}
  & \secondbest{17.52}
  & \secondbest{37.61}
  & 48.81 \\

\locar{} &
  & \secondbest{12.04}
  & 27.47
  & 37.53
  & 17.38
  & 37.49
  & \secondbest{48.91} \\

\locar{} + &
  & \best{12.71}
  & \best{28.70}
  & \secondbest{39.30}
  & \best{17.57}
  & \best{37.73}
  & \best{49.04} \\

\midrule

Zero-shot (Global) & \multirow{5}{*}{SigLIP2-B/16}
  & -- & -- & --
  & 22.80 & 43.58 & 53.76 \\

TextRegion &
  & 9.57 & 24.68 & 35.53
  & 24.69 & 46.66 & 57.63 \\

TextRegion + &
  & 10.10 & 26.40 & 37.70
  & 24.88 & 47.16 & 58.05 \\

\locar{} &
  & \secondbest{19.26}
  & \secondbest{39.19}
  & \secondbest{49.51}
  & \best{25.99}
  & \best{48.48}
  & \secondbest{59.54} \\

\locar{} + &
  & \best{19.95}
  & \best{40.63}
  & \best{51.42}
  & \secondbest{25.93}
  & \secondbest{48.45}
  & \best{59.70} \\

\midrule

Zero-shot (Global) & \multirow{3}{*}{Qwen3-VL-2B}
  & -- & -- & --
  & 20.17 & 40.56 & 51.92 \\

\locar{} &
  & \secondbest{19.90}
  & \secondbest{37.98}
  & \secondbest{47.80}
  & \secondbest{29.15}
  & \secondbest{52.60}
  & \secondbest{63.11} \\

\locar{} + &
  & \best{20.67}
  & \best{39.39}
  & \best{49.50}
  & \best{29.30}
  & \best{52.85}
  & \best{63.61} \\

\bottomrule
\end{tabular}
\end{table}

\textbf{Overall retrieval performance.}
Table~\ref{tab:main-results} shows that \locar{} + achieves the
highest fused recall among the evaluated configurations with Qwen:
$29.30/52.85/63.61\%$ at R@1/R@5/R@10.
Relative to the standalone global zero-shot rows, fused
\locar{} + improves these metrics by $2.42/4.90/6.15$ pp for CLIP,
$3.13/4.87/5.94$ pp for SigLIP2, and
$9.13/12.29/11.69$ pp for Qwen.
These comparisons establish gains over standalone retrieval,
but do not isolate fusion when the global configurations differ.
Against the matched global channels used in fusion, the R@1 gains
are $2.71$, $3.13$, and $2.82$ pp, respectively
(Appendix~\ref{app:zero-shot}).

\textbf{Region representation and paired support.}
Without paired support, \locar{} improves region R@1 over TextRegion
by $2.38$ pp for CLIP and $9.69$ pp for SigLIP2.
Adding support increases these total gains to $3.05$ and $10.38$ pp;
the residual itself contributes $0.67$ and $0.69$ pp.
Thus, the larger improvement comes from the canvas representation.
At R@10, \locar{} + exceeds unadapted TextRegion by $1.36$ pp
for CLIP and $15.89$ pp for SigLIP2.
The advantage is not uniform: CLIP TextRegion + retains a
$0.41$ pp lead over \locar{} + at region R@10.
For Qwen's crop representation, paired support adds
$0.77/1.41/1.70$ pp at R@1/R@5/R@10.



\textbf{Complementarity with global retrieval.}
Fused \locar{} + also outperforms fused TextRegion at all three
ranks: by $0.13/0.33/0.71$ pp for CLIP and
$1.24/1.79/2.07$ pp for SigLIP2.
The stronger region representation therefore contributes beyond
both global retrieval and the TextRegion baseline, particularly
with SigLIP2.
However, region improvements are attenuated by fusion.
For SigLIP2, adding support improves region recall at every rank
but slightly reduces fused R@1/R@5 by $0.06/0.03$ pp, while
increasing fused R@10 by $0.16$ pp.
Support improves all three fused metrics for all backbones.
Its benefit therefore depends on how the local scores complement
the global channel, rather than on region recall alone.


\textbf{Why do the gains differ across backbones?}
CLIP benefits from \locar{}, but its smaller region gains leave
less improvement to carry into fusion.
Canvas construction changes the visual input before encoding,
restricting attention to the crop and suppressing background detail.
TextRegion instead pools region features that already contain
context from the source image.
The results are consistent with canvas encoding providing more
discriminative local evidence, especially for SigLIP2.
However, this comparison jointly changes cropping, background
treatment, and feature aggregation; it does not isolate attention
locality or establish pretraining objectives as the cause of the
backbone differences.

\subsection{\locfg{}: transferred and in-domain support}
\label{sec:locfg}
\vspace{-10pt}

\begin{table}[t]
\centering
\small
\setlength{\tabcolsep}{2pt}
\renewcommand{\arraystretch}{1.08}

\caption{\textbf{LocFG retrieval} (\%).
``+'' adds paired support.
Tr and ID denote transferred and in-domain settings, including
their calibration choices.
Identical baseline values are shown once across both columns.
Separate baseline values reflect different fusion calibration
or, for Qwen, different query instructions.
Global zero-shot reports standalone retrieval; other right-hand
entries report global--region fusion.
Best and second-best values are marked within each backbone
and metric across both settings.
Qwen settings are detailed in Appendix~\ref{app:qwen}.}
\label{tab:locfg}
\label{tab:locfg-results}
\label{tab:full-locfg}
\label{tab:canvas-locfg-transfer}
\label{tab:canvas-locfg-indomain}
\label{tab:locfg-rep}

\begin{tabular*}{\linewidth}
  {@{\extracolsep{\fill}}l*{12}{r}@{}}
\toprule
& \multicolumn{6}{c}{\textbf{Region retrieval}}
& \multicolumn{6}{c}{\textbf{Global / fused}} \\
\cmidrule(lr){2-7}\cmidrule(l){8-13}

& \multicolumn{2}{c}{R@1}
& \multicolumn{2}{c}{R@5}
& \multicolumn{2}{c}{R@10}
& \multicolumn{2}{c}{R@1}
& \multicolumn{2}{c}{R@5}
& \multicolumn{2}{c}{R@10} \\
\cmidrule(lr){2-3}
\cmidrule(lr){4-5}
\cmidrule(lr){6-7}
\cmidrule(lr){8-9}
\cmidrule(lr){10-11}
\cmidrule(l){12-13}

\textbf{Configuration}
& Tr & ID & Tr & ID & Tr & ID
& Tr & ID & Tr & ID & Tr & ID \\
\midrule

\multicolumn{13}{@{}l}{\textbf{(a) CLIP-B/16}} \\
\addlinespace[2pt]

Global zero-shot
& \multicolumn{2}{c}{--}
& \multicolumn{2}{c}{--}
& \multicolumn{2}{c}{--}
& \multicolumn{2}{c}{7.01}
& \multicolumn{2}{c}{15.80}
& \multicolumn{2}{c}{20.76} \\
\addlinespace[3pt]

TextRegion
& \multicolumn{2}{c}{5.68}
& \multicolumn{2}{c}{13.71}
& \multicolumn{2}{c}{18.73}
& \multicolumn{2}{c}{8.23}
& \multicolumn{2}{c}{18.39}
& \multicolumn{2}{c}{24.76} \\

TextRegion +
& 5.76 & 5.80
& 13.79 & 13.84
& 18.79 & 18.86
& 8.30 & 8.30
& 18.42 & 18.42
& 24.79 & 24.79 \\
\addlinespace[3pt]

\locar{}
& \multicolumn{2}{c}{6.07}
& \multicolumn{2}{c}{14.42}
& \multicolumn{2}{c}{19.47}
& \multicolumn{2}{c}{\secondbest{8.44}}
& \multicolumn{2}{c}{\secondbest{18.81}} 
& \multicolumn{2}{c}{\secondbest{24.77}} \\

\locar{} +
& \secondbest{6.24} & \best{6.26}
& \secondbest{14.85} & \best{14.96}
& \secondbest{20.01} & \best{20.15}
& 8.41 & \best{8.76}
& 18.80 & \best{19.31}
& 24.68 & \best{25.85} \\

\midrule

\multicolumn{13}{@{}l}{\textbf{(b) SigLIP2-B/16}} \\
\addlinespace[2pt]

Global zero-shot
& \multicolumn{2}{c}{--}
& \multicolumn{2}{c}{--}
& \multicolumn{2}{c}{--}
& \multicolumn{2}{c}{10.20}
& \multicolumn{2}{c}{21.09}
& \multicolumn{2}{c}{27.32} \\
\addlinespace[3pt]

TextRegion
& \multicolumn{2}{c}{7.63}
& \multicolumn{2}{c}{17.50}
& \multicolumn{2}{c}{23.50}
& \multicolumn{2}{c}{12.00}
& \multicolumn{2}{c}{25.00}
& \multicolumn{2}{c}{32.00} \\

TextRegion +
& 7.71 & 7.75
& 17.58 & 17.53
& 23.62 & 23.79
& 12.05 & 12.12
& 25.05 & 25.05
& 32.05 & 32.05 \\
\addlinespace[3pt]

\locar{}
& \multicolumn{2}{c}{9.51}
& \multicolumn{2}{c}{20.84}
& \multicolumn{2}{c}{27.02}
& \multicolumn{2}{c}{\secondbest{12.04}}
& \multicolumn{2}{c}{\secondbest{25.45}}
& \multicolumn{2}{c}{\secondbest{33.11}} \\

\locar{} +
& \best{9.71} & \secondbest{9.57}
& \secondbest{21.33} & \best{21.41}
& \best{27.87} & \secondbest{27.61}
& 12.02 & \best{12.53}
& 25.19 & \best{26.56}
& 33.05 & \best{33.87} \\

\midrule

\multicolumn{13}{@{}l}{\textbf{(c) Qwen3-VL-2B}} \\
\addlinespace[2pt]

Global zero-shot
& \multicolumn{2}{c}{--}
& \multicolumn{2}{c}{--}
& \multicolumn{2}{c}{--}
& \multicolumn{2}{c}{11.78}
& \multicolumn{2}{c}{23.94}
& \multicolumn{2}{c}{31.16} \\
\addlinespace[3pt]

\locar{}
& \multicolumn{2}{c}{\secondbest{12.27}}
& \multicolumn{2}{c}{\secondbest{26.43}}
& \multicolumn{2}{c}{\secondbest{33.91}}
& \multicolumn{2}{c}{\secondbest{14.32}}
& \multicolumn{2}{c}{\secondbest{29.55}}
& \multicolumn{2}{c}{\secondbest{38.34}} \\

\locar{} +
& 11.08 & \best{12.46}
& 24.05 & \best{26.71}
& 31.53 & \best{34.26}
& 13.73 & \best{14.40}
& 28.31 & \best{29.60}
& 37.09 & \best{38.38} \\

\bottomrule
\end{tabular*}
\end{table}


We evaluate generalization on \locfg{}, our small-referent retrieval
benchmark introduced in Sec.~\ref{sec:setup}.
All results in Table~\ref{tab:locfg-results} use the same \locfg{}
test queries and gallery.
The transferred setting (Tr) reuses the RefCOCOg support bank and
calibration, whereas the in-domain setting (ID) uses support and
calibration from the image-disjoint \locfg{} fit partition.
This comparison tests whether support from another dataset remains
useful and whether adapting to the target domain provides additional
benefit.

Rows without ``+'' are unadapted configurations: they use no
paired-support residual, corresponding to $\gamma=0$.
Rows with ``+'' add the paired expression--region residual.
For a fixed representation and query instruction, unadapted region
scores are independent of the support bank.
Fused scores can nevertheless differ across settings because the calibrated fusion weight changes.


The three backbone panels include standalone global retrieval
and \locar{} with and without paired support.
CLIP and SigLIP2 additionally include TextRegion and TextRegion +.
Qwen uses plain proposal crops, and its transferred and in-domain
runs use different query instructions.
Consequently, the Qwen comparison reflects changes in both
instruction and support configuration
(Appendix~\ref{app:qwen}).




\textbf{Overall retrieval performance.}
The strongest reported \locfg{} configuration is in-domain
Qwen \locar{} +, reaching region recall of
$12.46/26.71/34.26\%$ and fused recall of
$14.40/29.60/38.38\%$ at R@1/R@5/R@10.
Relative to the standalone global rows, in-domain fused
\locar{} + improves these metrics by $1.75/3.51/5.09$ pp
for CLIP, $2.33/5.47/6.55$ pp for SigLIP2, and
$2.62/5.66/7.22$ pp for Qwen.
Thus, the complete configurations improve retrieval beyond
standalone global embeddings on all three backbones.
For Qwen, this standalone comparison also includes differences
in the query configuration; matched global comparisons are
reported in Appendix~\ref{app:qwen}.

\textbf{Representation gains and transferred support.}
Without paired support, \locar{} improves region R@1/R@10 over
TextRegion by $0.39/0.74$ pp for CLIP and $1.88/3.52$ pp for
SigLIP2.
Transferred \locar{} + increases these total gains to
$0.56/1.28$ and $2.08/4.37$ pp, respectively.
The transferred residual therefore improves region recall at
all three ranks for these two backbones.
However, its fused R@1/R@10 effects are slightly negative:
$-0.03/-0.09$ pp for CLIP and $-0.02/-0.06$ pp for SigLIP2.
For Qwen, transferred support raises region R@1 by $0.12$ pp
but lowers R@5 by $0.08$ pp.
Transferred support can improve local matching, but its benefit
is neither uniform across metrics nor guaranteed to improve fusion.

\textbf{What does in-domain support add?}
In-domain \locar{} + exceeds TextRegion at region R@1/R@10
by $0.58/1.42$ pp for CLIP and $1.94/4.11$ pp for SigLIP2.
Relative to \locar{} without support, the region R@1 increments
are modest: $0.19$ pp for CLIP, $0.06$ pp for SigLIP2, and
$0.19$ pp for Qwen.
In-domain \locar{} + achieves higher fused recall than its
transferred counterpart at all three ranks for every backbone.
This does not establish that the support bank alone is responsible:
fusion calibration changes, and Qwen additionally changes its
query instruction.
Nor is in-domain support uniformly stronger for region retrieval:
SigLIP2's transferred \locar{} + retains higher R@1 and R@10
than its in-domain counterpart.

\textbf{Interpreting the smaller gains.}
The smaller CLIP and SigLIP2 region improvements than on RefCOCOg
show that the benefit of canvas encoding depends on the evaluation
setting.
LocFG emphasizes smaller referents: the median annotated box
occupies $3.0\%$ of image area, compared with $14.5\%$ on RefCOCOg.
At our operating point, only $51.1\%$ of LocFG queries have a
retained proposal with box IoU $\geq 0.5$ against the referent.
Limited availability of well-localized candidates may therefore
constrain region encoding, although this coverage is not an upper
bound on image-retrieval recall.
Together, the results motivate better small-region proposals and
representations; they do not isolate object size, proposal quality,
or pretraining as the sole cause of the cross-dataset differences.

\subsection{Ablation studies}
\label{sec:ablations}
\label{sec:canvas-ablation}
\label{sec:tipregion}
\vspace{-10pt}

In this section we examine  various canvas masking, pooling startegies, text instructions template and cache capacity usage. Details on calibration settings, cache sizes and Qwen instruction selection  are in
Appendix ~\ref{app:eval-protocol}, ~\ref{app:locar-cache}, ~\ref{app:qwen}. 

\textbf{Background masking} \locar{} constructs region embeddings by independently encoding each SAM2 proposal as a canvas. The foreground region is preserved but the surrounding background is suppressed and the composite image is passed through the frozen image encoder. A key design choice is how aggressively to suppress the background (a) soft suppression via Gaussian blur keeps the surround visually plausible and in-distribution for the encoder (b) while hard suppression via white fill removes all contextual information at the cost of introducing an out-of-distribution background. We ablate these two treatments on SigLIP2-B/16 with the RefCOCOg test split as shown in \ref{tab:canvas-design}. White fill reduces region R@1 by 2.72 pp versus blur but improves fused R@1 by 0.30 pp, likely because stronger background removal makes region errors more complementary to the global embedding. Blur preserves more context and gives stronger standalone region representations, while white fill provides slightly better fusion. Since blur still outperforms TextRegion by +9.69 pp region R@1 and +1.30 pp fused R@1, we use blur as the default; the white-fill result mainly highlights the tradeoff between region quality and global-region complementarity.

\textbf{Proposal vector pooling}
After encoding each canvas crop with the frozen ViT, we compare two region readouts in ~\ref{tab:canvas-mean-pooling} the attention-pooled CLS representation and mean pooling over patch tokens. Mean pooling performs substantially worse: region R@1 improves by only +0.75 pp over TextRegion, compared with +9.69 pp using CLS, while fused R@1 falls below TextRegion (23.67\% vs. 24.69\%). A likely reason is that the patch tokens are intermediate visual features and are not individually optimized to align with text. Simply averaging them therefore does not necessarily produce a meaningful point in the joint image–text embedding space. In contrast, the CLS readout is explicitly optimized through the model's contrastive objective to represent the image in that space. Paired support adds only +0.30 pp with mean pooling, further suggesting that support cannot fully compensate for a poorly aligned base representation. Overall, this ablation shows that how the canvas is pooled is critical: independent encoding is most effective when its text-aligned CLS readout is preserved.
\begin{table}[htbp]
\centering
\normalsize
\setlength{\tabcolsep}{3pt}
\renewcommand{\arraystretch}{1.05}

\caption{\textbf{SigLIP2 background ablation}
on RefCOCOg (\%).
Both configurations use the default attention-pooling readout,
labeled \texttt{cls} in the experiment logs.
The shared global channel gives $22.80/43.58/53.76$
at R@1/R@5/R@10.}
\label{tab:canvas-design}
\label{tab:canvas-background}

\begin{tabular}{@{}llrrrrrr@{}}
\toprule
& & \multicolumn{3}{c}{\textbf{Region}}
  & \multicolumn{3}{c}{\textbf{Fused}} \\
\cmidrule(lr){3-5}\cmidrule(l){6-8}
Background & Support
  & R@1 & R@5 & R@10
  & R@1 & R@5 & R@10 \\
\midrule

\multirow{2}{*}{Blur}
& Without
  & 19.26 & 39.19 & 49.51
  & 25.99 & 48.48 & 59.54 \\
& With
  & 19.95 & 40.63 & 51.42
  & 25.93 & 48.45 & 59.70 \\
\midrule

\multirow{2}{*}{White}
& Without
  & 16.54 & 33.49 & 43.43
  & 26.29 & 49.10 & 60.06 \\
& With
  & 17.38 & 35.75 & 46.35
  & 26.17 & 49.11 & 60.30 \\

\bottomrule
\end{tabular}
\end{table}

\begin{table*}[htbp]
\centering
\normalsize
\setlength{\tabcolsep}{3pt}
\renewcommand{\arraystretch}{1.05}

\caption{\textbf{SigLIP2 mean-pooling ablation}
on RefCOCOg (\%).
Region canvases use blurred backgrounds and mean pooling.
Support denotes the paired-support residual.
The global channel is identical to that in
Table~\ref{tab:canvas-background}.}
\label{tab:canvas-mean-pooling}

\begin{tabular}{@{}lrrrrrr@{}}
\toprule
& \multicolumn{3}{c}{\textbf{Region}}
& \multicolumn{3}{c}{\textbf{Fused}} \\
\cmidrule(lr){2-4}\cmidrule(l){5-7}
Support
  & R@1 & R@5 & R@10
  & R@1 & R@5 & R@10 \\
\midrule

Without
  & 10.32 & 26.23 & 36.63
  & 23.67 & 45.41 & 56.01 \\
With
  & 10.62 & 26.60 & 36.93
  & 23.72 & 45.46 & 56.08 \\

\bottomrule
\end{tabular}
\end{table*}

\section{Conclusion}
\label{sec:conclusion}
\vspace{-10pt}
We presented \locar{}, a training-free approach that independently
encodes proposal-based region canvases with frozen vision--language
models and combines local evidence with whole-image similarity.
Experiments on RefCOCOg and our small-referent benchmark \locfg{}
show that improving region representations provides larger gains
than the additional paired-support residual.
Across the backbones \locar{} + shows lift of 2.40-9.13 pp on recall metrics on RefCOCOg.
Ablations highlight the importance of background treatment and
visual readout.
Future work will focus on better small-referent proposals,
support adaptation under domain shift, and more efficient
region encoding and caching.

\clearpage
\section*{AI Use Statement}
Generative AI tools were used for language editing, LaTeX restructuring, and the initial layout of explanatory figures. All scientific claims, equations, experimental results, statistical tests, and bibliographic entries were checked by the authors against the experiment artifacts and cited sources. No AI system selected experimental outcomes or replaced human verification of the reported numbers.

\bibliographystyle{iclr2027_conference}
\bibliography{references}

\clearpage
\appendix
\providecommand{\locfg}{\textsc{LocFG}}

\section{Extended evaluation and ablations}
\label{app:eval-details}

\subsection{Implementation, support filtering, and metrics}
\label{app:eval-protocol}

\paragraph{Data and support separation.}
RefCOCOg test contains $9{,}602$ expressions over $2{,}600$ gallery
images. Removing $481$ validation rows shared with test leaves
$6{,}843$ support candidates. We retain up to $4{,}096$ matched
expression--region pairs and audit each bank for evaluation overlap.
Hyperparameters are selected on held-out support images before
test evaluation. Transferred evaluation reuses the RefCOCOg bank
and calibration; in-domain \locfg{} evaluation uses its fit partition.

\locfg{} is derived from Visual Genome~\citep{krishna2017visualgenome}
and contains $12{,}033$ test queries over $2{,}600$ images.
Its fit/test partitions are image-disjoint, and a perceptual-hash
blocklist filters overlap with RefCOCOg..

\paragraph{Referent size and proposal coverage.}
The median referent occupies $3.0\%$ of image area on \locfg{},
compared with $14.5\%$ on RefCOCOg.
At the evaluated proposal settings, $51.1\%$ of \locfg{} queries
have a proposal with box IoU $\geq 0.5$ against the annotated
referent. This measures proposal localization coverage, not an
upper bound on source-image retrieval recall.
The coverage-stratified experiments do not establish coverage
as the sole explanation for the small support gains.

\paragraph{Proposals and matched comparisons.}
SAM2 \texttt{hiera-large} uses $16$ points per side, and scoring
retains at most the first $32$ proposals per image.
Within CLIP and SigLIP2, representation comparisons share proposal
identities and ordering, query embeddings, and whole-image embeddings.
TextRegion provides a frozen mask-pooled comparator; it is not
required to construct or score \locar{} canvases.
Qwen independently encodes proposal crops because its interface
does not expose the patch-to-text projection used by our
TextRegion implementation.

\paragraph{Canvas construction.}
\label{app:canvas-construction}
Each proposal box is expanded by $10\%$ on each side.
The canvas preserves pixels inside the proposal mask and blurs
the surrounding background with a $55\times55$ Gaussian kernel.
The composite is resized to $224\times224$ and independently
encoded by the frozen CLIP or SigLIP2 image encoder.
White-fill controls replace the background with white.
The default readout is labeled \texttt{cls} in the experiment
logs; this label does not imply identical pooling architectures
across backbones. The mean variant uses mean pooling.
Comparing canvases with TextRegion jointly changes cropping,
resizing, background treatment, and independent encoding.

\paragraph{Scoring and calibration.}
For normalized query and region embeddings $t$ and $r$, the
unadapted region score is
\[
s_0(T,I)=\max_{r\in\mathcal{R}(I)}\langle t,r\rangle.
\]
The paired-support residual compares query affinities to support
expressions with region affinities to their paired visual anchors,
yielding $s_\gamma$. Setting $\gamma=0$ recovers $s_0$.
For a normalized whole-image embedding $g(I)$,
$s_{\mathrm{glob}}(T,I)=\langle t,g(I)\rangle$, and fusion uses
\[
s_{\mathrm{fuse}}(T,I)
=
z_T[s_{\mathrm{glob}}](I)
+\alpha z_T[s_\gamma](I).
\]
For each query and channel, $z_T$ subtracts the gallery mean
and divides by the gallery standard deviation.
Unadapted region scores require no support, whereas unadapted
fused scores use a calibrated $\alpha$.
Matched comparisons without and with support share $\alpha$
within each representation and setting.
On \locfg{}, SigLIP2 canvas fusion uses $\alpha=0.5$ under
transfer and $\alpha=1.0$ in-domain; cross-setting fused
comparisons therefore include calibration changes.

\subsection{Global baselines and fusion gains}
\label{app:zero-shot}
\label{app:full-recall}

Tables~\ref{tab:main-results} and~\ref{tab:locfg-results} report
the main retrieval comparisons.
Table~\ref{tab:zero-shot} collects standalone global results.
These differ from the matched global channels used in the
RefCOCOg fusion runs, whose R@1 values are $14.86$, $22.80$,
and $26.48$ for CLIP, SigLIP2, and Qwen, respectively.
Subtracting a standalone global result from a fused result does
not isolate fusion unless prompts, preprocessing, and scoring
are matched.

\begin{table}[t]
\centering
\normalsize
\caption{\textbf{Standalone global zero-shot retrieval} (\%).}
\label{tab:zero-shot}
\setlength{\tabcolsep}{4pt}
\begin{tabular*}{\linewidth}
  {@{\extracolsep{\fill}}llrrr@{}}
\toprule
Dataset & Backbone & R@1 & R@5 & R@10 \\
\midrule
RefCOCOg & CLIP-B/16     & 15.15 & 32.83 & 42.89 \\
         & SigLIP2-B/16  & 22.80 & 43.58 & 53.76 \\
         & Qwen3-VL-2B   & 20.17 & 40.56 & 51.92 \\
\midrule
\locfg{} & CLIP-B/16     & 7.01  & 15.80 & 20.76 \\
         & SigLIP2-B/16  & 10.20 & 21.09 & 27.32 \\
         & Qwen3-VL-2B   & 11.78 & 23.94 & 31.16 \\
\bottomrule
\end{tabular*}
\end{table}

\paragraph{Relative gains and backbone comparisons.}
On RefCOCOg, fused \locar{} + exceeds matched global R@1 by
$2.71$ pp for CLIP, $3.13$ pp for SigLIP2, and $2.82$ pp for Qwen,
corresponding to relative gains of $18.2\%$, $13.7\%$, and $10.6\%$.
The support-only region R@1 gain is $0.67/12.04=5.6\%$ for CLIP
and $0.77/19.90=3.9\%$ for Qwen.
Their fused R@10 increments are $0.13$ and $0.50$ pp, respectively.
Thus, Qwen's larger fused R@10 increment does not imply a uniformly
larger relative region improvement.
Aggregate recalls alone establish neither complementary error
patterns nor a statistical ordering across backbones.

\subsection{Canvas background and pooling ablations}
\label{app:canvas-ablations}

Tables~\ref{tab:canvas-design} and~\ref{tab:canvas-mean-pooling}
compare background treatment and readout for SigLIP2-B/16 on
RefCOCOg.
Blur with the default readout uses
$\gamma=0.10$, $K=64$, and $\alpha=0.5$;
white uses $\gamma=0.10$, $K=32$, and $\alpha=0.5$;
blur with mean pooling uses
$\gamma=0.05$, $K=64$, and $\alpha=0.25$.
Unadapted rows set $\gamma=0$.
Unadapted region comparisons therefore exclude support and fusion,
whereas adapted and fused comparisons include these calibrated settings.

\paragraph{Background treatment.}
Without support, blur exceeds white fill in region R@1/R@5/R@10
by $2.72/5.70/6.08$ pp.
With support, the differences are $2.57/4.88/5.07$ pp.
However, white achieves higher fused R@1:
$26.29$ versus $25.99$ without support and
$26.17$ versus $25.93$ with support.
The reported McNemar values ($p=0.53$ for white and $p=0.71$
for blur) compare predictions without and with support within
each setting. They do not test white against blur or establish
equivalence between backgrounds.

\paragraph{Readout and fusion.}
Under blur, replacing the default readout with mean pooling
reduces unadapted region R@1 by $8.94$ pp, from $19.26$ to $10.32$.
Adapted fused R@1 is also lower, at $23.72$ versus $25.93$,
although this comparison changes $\gamma$ and $\alpha$ as well
as the readout.
Z-score standardization removes channel offsets and scales;
it does not improve within-channel rankings.
Aggregate recalls alone do not establish whether similar fused
scores arise from complementary errors or dominance of the
global channel.

\subsection{Qwen crops and instruction selection on \locfg{}}
\label{app:qwen}

The incumbent instruction (I) is \texttt{asym-image}, and the
fit-selected instruction (S) is \texttt{asym-scene-part}.
The instruction comparison fixes visual features and uses
$\gamma=0.05$, $K=16$, and $\alpha=1$.
Instructions are selected using fit data.
Table~\ref{tab:qwen-locfg} reports the original instruction-ablation
results, with unadapted region recalls, support increments,
and matched global and fused R@1.

\begin{table}[t]
\centering
\normalsize
\caption{\textbf{Qwen instruction comparison on \locfg{}.}
Region triplets are R@1/R@5/R@10 (\%); gains are percentage points.
Global and fused columns report matched unadapted R@1.}
\label{tab:qwen-locfg}
\label{tab:locfg-qwen}
\setlength{\tabcolsep}{3pt}
\begin{tabular*}{\linewidth}
  {@{\extracolsep{\fill}}lcccc@{}}
\toprule
Instruction & Region recall & Support gain
& Global R@1 & Fused R@1 \\
\midrule
Incumbent (I)
& 10.96/24.13/31.52
& +0.23/+0.43/+0.52
& 10.90 & 13.64 \\
Selected (S)
& 12.27/26.43/33.91
& +0.18/+0.28/+0.36
& 11.70 & 14.32 \\
\bottomrule
\end{tabular*}
\end{table}

Selecting S improves unadapted region R@1/R@5/R@10 by
$1.31/2.30/2.39$ pp, matched global R@1 by $0.80$ pp,
and unadapted fused R@1 by $0.68$ pp.
The incremental support gains decrease by $0.05/0.15/0.16$ pp.
Thus, S improves region retrieval while leaving a smaller
additional gain from paired support in this comparison.
The matched global baselines differ from the standalone
Qwen zero-shot R@1 of $11.78$.

\begin{table}[t]
\centering
\normalsize
\caption{\textbf{Qwen instruction slate on \locfg{}.}
Unadapted region recalls (\%) and paired-support gains (pp).
Instructions are selected on fit data.}
\label{tab:qwen-instruction-slate}
\setlength{\tabcolsep}{3pt}
\begin{tabular*}{\linewidth}
  {@{\extracolsep{\fill}}lccc@{}}
\toprule
Instruction & $\beta$
& R@1/R@5/R@10 & Gain R@1/R@5/R@10 \\
\midrule
Model default
& 12.72 & 8.22/18.06/23.73 & +0.14/+0.91/+1.30 \\
\texttt{asym-object}
& 11.84 & 10.08/21.21/27.65 & +0.11/+0.73/+0.85 \\
\texttt{asym-region}
& 12.09 & 11.34/24.15/31.45 & +0.08/+0.38/+0.53 \\
\texttt{asym-image} (I)
& 11.40 & 10.96/24.13/31.52 & +0.23/+0.43/+0.52 \\
\texttt{asym-scene-part} (S)
& 12.24 & 12.27/26.43/33.91 & +0.18/+0.28/+0.36 \\
\bottomrule
\end{tabular*}
\end{table}

Image-cluster sign-flip tests give R@10 $p<10^{-4}$ for
model-default and object instructions, $p=0.0001$ for region
and image instructions, and $p=0.0058$ for the scene-part instruction.
The corresponding cluster-bootstrap $95\%$ gain intervals are
$[1.01,1.58]$, $[0.57,1.13]$, $[0.27,0.80]$,
$[0.28,0.78]$, and $[0.11,0.61]$ pp.
These tests assess support gains within each instruction,
not differences between instructions.

\subsection{Cache storage}
\label{app:locar-cache}

Table~\ref{tab:locar-cache-size} reports the saved gallery-cache
and support-bank sizes for each backbone.
Their combined storage ranges from $133$ to $298$ MB.
These figures describe stored files and do not measure peak
memory consumption during scoring.

\begin{table}[t]
\centering
\small
\setlength{\tabcolsep}{8pt}
\renewcommand{\arraystretch}{1.05}
\caption{\textbf{\locar{} cache storage} (MB).
Combined denotes the sum of the reported gallery cache
and support bank.}
\label{tab:locar-cache-size}

\begin{tabular}{@{}lrrr@{}}
\toprule
Backbone & Gallery cache & Support bank & Combined \\
\midrule
CLIP-B/16   & 78  & 55 & 133 \\
SigLIP2-B/16 & 116 & 80 & 196 \\
Qwen3-VL-2B & 270 & 28 & 298 \\
\bottomrule
\end{tabular}
\end{table}

\section{Support-set contamination audit}
\label{sec:contamination}

\paragraph{Split-resolution failure.}
The public RefCOCOg mirror used in our earlier pipeline contains
only \texttt{val-*} and \texttt{test-*} parquet shards.
When requested to load the missing \texttt{train} split, a
file-globbing fallback returns the union of validation and test.
Separately, validation and test contain $481$ shared rows with
identical file and question identifiers.
The declared split, bank size, and feature statistics remain
plausible, so these checks alone do not establish support provenance.

\paragraph{Paired feature audit.}
Support anchors are raw features from the same frozen encoder
used for evaluation.
For each evaluation pair, we identify its nearest text anchor
and nearest region anchor.
We flag potential reuse only when both cosine similarities
exceed $\tau$ and the two nearest-anchor indices agree.
Requiring a shared index distinguishes paired overlap from
repeated or formulaic language.

On the clean bank, lowering $\tau$ from $0.9999$ to $0.95$
increases text-only matches from $1.63\%$ to $70.93\%$,
while paired matches increase from $0.00\%$ to $0.03\%$.
The contaminated bank remains near $17\%$ paired overlap
throughout the sweep.
Table~\ref{tab:audit-main} reports the strict-threshold results.

\begin{table}[t]
\centering
\small
\caption{\textbf{Support audit at $\tau=0.9999$}
on $3{,}000$ RefCOCOg test queries.
Paired overlap requires matching text and region features at
the same support index.
The $481$ shared rows must be removed even when validation
is loaded correctly.}
\label{tab:audit-main}
\setlength{\tabcolsep}{4pt}
\begin{tabular}{@{}lcc@{}}
\toprule
Support bank & Declared source & Paired overlap \\
\midrule
All shards via missing-split fallback
& \texttt{train} & $16.93\%$ \\
Unfiltered validation
& \texttt{val} & $2.60\%$ \\
Validation with shared test rows removed
& \texttt{val} & $0.00\%$ \\
\bottomrule
\end{tabular}
\end{table}

\paragraph{Effect on retrieval.}
The contaminated scorer achieves $59.3\%$ R@1 on the $508$
flagged queries, compared with $5.42\%$ $[4.6,6.4]$ on the
remaining $2{,}492$.
The much higher recall among flagged queries is consistent
with support reuse inflating retrieval performance.
A two-bank mixture calculation agrees on the clean rate but
estimates a higher leaked rate because it assumes identical
conditional behavior across banks.
We retain that calculation as a cross-check and prioritize the direct
conditional measurements.
RefCOCOg support banks are constructed from filtered validation
data and must pass the audit before scoring.
\section{\textsc{LocFG} construction}
\label{app:locfg-stress}
\paragraph{Construction.} From Visual Genome~\citep{krishna2017visualgenome} we emit two splits over
shared images. \emph{Retrieval}: a region description and the box it describes, at most $8$ regions
per image, phrases of at least three content words, deduplicated within an image. \emph{Compositional}:
image versus \{correct, hard-negative\} caption in three grounded categories --- \texttt{attribute}
(within-type attribute swap), \texttt{relation} (subject/object order swap of a directional
predicate) and \texttt{object} (plausible absent same-frequency object replacement) --- at most two
items per category per image. Negatives are generated with length- and frequency-matching
safeguards so a bag-of-words model cannot win by surface statistics; the length-bias check and a
$50$-per-category human-verification sample ship with the release.

\paragraph{Leakage control, which is the point.} Images are filtered against RefCOCOg and the
compositional benchmarks by a dHash plus banded-LSH blocklist, catching both COCO-identifier
matches and perceptual near-duplicates. The blocklist is released alongside the data, because
``leakage-controlled'' is otherwise an unverifiable claim. \texttt{locfg\_split.py} then carves an
image-disjoint \texttt{fit}/\texttt{test} partition by a stable hash of the image identifier, so a
method may be adapted in-domain on \texttt{fit} and evaluated on \texttt{test} without overlap.
\paragraph{A caveat we measured rather than inherited.} Small referents are hard for
class-agnostic proposers, and this bounds what any proposal-scoring method can do here. SAM2 at
the settings TextRegion uses covers \textsc{LocFG} boxes poorly. Over all $12{,}033$ evaluated
queries, only $51.1\%$ have a proposal reaching IoU $0.5$ with the ground-truth box inside the
$R_{\max}{=}32$ scoring cap (median $0.512$), against $85.8\%$ on RefCOCOg, and $16.8\%$ have no
proposal above IoU $0.1$ at all. Lifting the cap recovers little --- $54.2\%$ at IoU $0.5$
uncapped, median $0.549$, $14.8\%$ still unmatchable, saturating by a cap of $64$ --- so the
referents are mostly not proposed, rather than proposed and then discarded. Users of \textsc{LocFG} who score
proposals rather than ground-truth boxes should measure and report this; it is a property of the
segmenter meeting small referents, not of the annotations.

\section{\textsc{LocFG}: eliminations in full}
\label{app:locfg}
\paragraph{No $\gamma$ rescues it, and the failure is domain-specific.} Because our own
calibration could not resolve $\gamma$ , a null at one value would be weak
evidence, so we swept the whole band and one value beyond it (Table~\ref{tab:locfg}). The transfer
is null at every $\gamma$ the support set left standing, and the trend runs \emph{opposite} to
RefCOCOg's. Beyond the band the two diverge sharply: at $\gamma{=}0.30$ the residual \emph{costs}
\textsc{LocFG} $-1.94$ R@10 ($p{=}3.3\times10^{-13}$) while still \emph{gaining} $+1.86$ R@10 on
RefCOCOg ($p{=}8.3\times10^{-7}$) with the identical bank and backbone. Upweighting the residual is
therefore not generically harmful; it is harmful only where the exemplar structure does not apply,
which is what an inert term would not do. The RefCOCOg column is the control that licenses this
reading, and without it the out-of-band row would have been uninterpretable.

\begin{table}[t]
\centering
\caption{\textbf{\textsc{LocFG} transfer, $\gamma$ sweep}, RefCOCOg-built anchors applied unchanged.
$\Delta$ against $\gamma{=}0$ on the region channel; $2{,}600$ images / $12{,}033$ queries. The
RefCOCOg column is the same backbone and bank in-domain, at the same $\gamma$, as a control. Every
\textsc{LocFG} entry inside the calibrated band $[0.05, 0.15]$ is non-significant; the out-of-band
row separates the two datasets in \emph{opposite} directions, both significantly.}
\label{tab:locfg}
\small
\begin{tabular}{l r r r l r}
\toprule
 & \multicolumn{4}{c}{\textsc{LocFG} (transferred)} & RefCOCOg \\
$\gamma$ & $\Delta$R@1 & $\Delta$R@5 & $\Delta$R@10 & $p$ (R@1 / R@10) & $\Delta$R@10 \\
\midrule
$0.05$          & $+0.12$ & $+0.02$ & $+0.17$ & $0.24$ / $0.19$ & $+1.24$ \\
$0.10^{\ast}$   & $+0.08$ & $+0.08$ & $+0.12$ & $0.59$ / $0.50$ & $+2.17$ \\
$0.15$          & $-0.03$ & $-0.09$ & $-0.16$ & $0.88$ / $0.45$ & $+2.59$ \\
$0.30$ (out of band) & $-0.61$ & $-1.97$ & $-1.94$ & $\mathbf{2.8\!\times\!10^{-3}}$ / $\mathbf{3.3\!\times\!10^{-13}}$ & $\mathbf{+1.86}$ \\
\bottomrule
\end{tabular}
\end{table}
\paragraph{Coverage is necessary but not sufficient, and we tested that rather than assuming it.}
The obvious reconciliation of \textsc{LocFG}'s flat profile is that its ``well-covered'' half is
well covered only by its own standard --- $51\%$ of its queries clear IoU $0.5$ against RefCOCOg's
$86\%$ (both over all evaluated queries at the $R_{\max}{=}32$ cap), its conditional anchor median
$0.586$ against $0.883$ --- so it may simply never reach the regime in
which the depth gain exists. That predicts a depth gain on the \textsc{LocFG} queries that
\emph{do} reach RefCOCOg's regime. It is false. Splitting instead at IoU $0.8$ isolates $2{,}542$
\textsc{LocFG} queries whose referent SAM2 covers as well as a typical RefCOCOg one, and they show
$+0.43 / +0.12 / +0.12$, with a top-minus-bottom contrast of $+0.40$, $+0.11$ and $-0.22$, none
significant ($p{=}0.31$, $0.82$, $0.65$). Restoring coverage on \textsc{LocFG} does not restore the
effect.

We therefore report coverage as \emph{necessary and not sufficient}: within RefCOCOg, losing
coverage removes the deep-rank gain; on \textsc{LocFG}, having coverage does not produce it. What
distinguishes the two beyond coverage is not established here, and we flag one confound in our own
test rather than leave it for a reader to find. The stratification controls coverage on the
\emph{query} side only; the \textsc{LocFG} bank itself was built from anchors with median IoU
$0.586$, so even a well-covered query is matched against noisy exemplars. We tested that: rebuilding
the bank at \texttt{--iou-min 0.7} yields $2{,}827$ anchors at median IoU $0.836$, comparable to
RefCOCOg's $0.883$, and the aggregate does not improve ($-0.02$ R@1; the small decline is within
what the scaling curve attributes to the smaller bank). Bank quality is therefore not the
obstacle either.

What replicates across both banks is narrower and stranger. The R@1 coverage contrast is $+0.76$
($p{=}8.2\times10^{-3}$) with the unfiltered bank and $+0.75$ ($p{=}8.4\times10^{-3}$) with the
filtered one --- two independently constructed banks agreeing to the second decimal --- while the
R@10 contrast is $+0.30$ and $+0.06$, neither significant. \textsc{LocFG} thus has a small,
coverage-gated rank-$1$ effect and no depth effect at all, which is the mirror image of RefCOCOg,
where coverage gates the depth effect ($+1.89$, $p{=}0.021$) and leaves rank-$1$ untouched
($-0.23$, $p{=}0.71$). We do not have an account of why the two invert, and we prefer to report the
inversion than to supply one after the fact; two of our earlier explanations for this benchmark
were falsified by the next control we ran.

A third candidate is also eliminated by measurement: that string-unique VG phrases describe
visually \emph{commonplace} referents, so deep-rank promotion is spread across lookalike regions
in other images and nets nothing. Counting gallery images that contain a near-copy of each query's
referent (region-feature cosine $\ge 0.9$) gives a \emph{median of zero} on \textsc{LocFG}; the
unique-vs-most-duplicated contrast is flat at every level (R@10 $+0.34$, $p{=}0.58$); and RefCOCOg
in fact contains \emph{more} such duplication (median $4$ vs $2$ at cosine $\ge 0.85$) while
keeping its depth gains. Whatever suppresses the depth component here, it is not that the
referents are commonplace, and it is not proposal coverage, and it is not bank quality.

The last candidate we could name was the query distribution itself: RefCOCOg expressions are
\emph{discriminative} by construction, Visual Genome descriptions merely describe, and a mechanism
that matches paired exemplars might need the contrastive structure. The one operational form we
could derive from that conjecture is profile collapse --- descriptive queries selecting the same
few anchors, driving the residual toward the query-\emph{independent} per-image score that the
region-marginal control already measures as inert. Measured, the collapse is absent, and absent on
both datasets alike: queries use $1{,}676$ of $4{,}096$ anchors effectively on \textsc{LocFG}
against $1{,}923$ on RefCOCOg, mean pairwise profile cosine is $0.008$ on both, and the mean
correlation between residual score rows is $0.037$ against $0.034$. The residual is as
query-dependent on the benchmark where it does nothing as on the benchmark where it works.
\paragraph{An eval-read ledger for this benchmark.} The experiments above read \textsc{LocFG}
test under six scoring configurations: the pre-registered transfer ($\gamma{=}0.10$, one read),
three band-sensitivity values ($\gamma \in \{0.05, 0.15, 0.30\}$, labelled analysis), the
in-domain bank at transferred hyperparameters, and the IoU-filtered in-domain bank. The
stratifications (coverage at two thresholds, and the diagnostics below) re-partition those same
scoring passes rather than scoring new configurations, and select nothing. We state this because
the count is high enough that a reader should get to audit it rather than trust it: none of the
six was chosen on the basis of a previous read's outcome except as disclosed retractions in this
section, and any future configuration on this benchmark --- a support-selected region cap, a
well-posedness query filter, an in-domain calibration --- will be frozen on the fit split and
read exactly once, extending this ledger.

\end{document}